\documentclass{amsart}

\usepackage[english]{babel}

\usepackage[letterpaper,top=2cm,bottom=2cm,left=3cm,right=3cm,marginparwidth=1.75cm]{geometry}

\usepackage{amssymb}
\usepackage{amsmath}
\usepackage{amsthm}
\usepackage{lipsum}
\usepackage{color}
\usepackage{enumerate}
\usepackage{caption}
\usepackage[labelformat=simple]{subcaption}

\usepackage{tikz}
\usetikzlibrary{graphs, graphs.standard,fit,positioning}
\usepackage{lipsum}

\newtheorem{theorem}{Theorem}
\newtheorem{lemma}[theorem]{Lemma}

\newtheorem{definition}[theorem]{Definition}

\newtheorem{aragao}[theorem]{Example}
\newtheorem{aragaoProva}[theorem]{Proof}

\title{Imprecise Belief Fusion Facing a DST Benchmark Problem}

\address{Federal University of Ceará, Campus Quixadá, Quixadá, Ceará, Brazil}
\author{F.~Aragão}
\email{aragao@ufc.br} 

\address{Federal University of Ceará, Campus Fortaleza, Quixadá, Ceará, Brazil}
\author{J. Alcântara}
\email{jnandoalc@ufc.br}

\begin{document}

\begin{abstract}

When we merge information in Dempster-Shafer Theory (DST), we are faced with anomalous behavior: agents with equal expertise and credibility can have their opinion disregarded after resorting to the belief combination rule of this theory. This problem is interesting because belief fusion is an inherent part of dealing with situations where available information is imprecise, as often occurs in Artificial Intelligence. We managed to identify an isomorphism 
betwin the DST formal apparatus into that of a Probabilistic Logic. 
Thus, we solved the problematic inputs affair by replacing the DST combination rule with a new fusion process aiming at eliminating anomalies proposed by that rule. 
We apply the new fusion method to the DST paradox Problem.

\end{abstract}

 \maketitle
 
\section{Introduction}
\label{sec:introduction}

In Dempster Shafer Theory (DST), 
personal opinions not empirically proven (called beliefs) are formalized. It offers a way to put together the information obtained from different independent sources, combining them into a new resulting one. This process is known as Belief Fusion and  
was introduced by Shafer in \cite{shafer1976mathematical} 
and stemmed from Dempster’s works \cite{10.1214/aoms/1177698950, 10.2307/2984504}.
DST has been employed as a technique for modelling reasoning under uncertainty.

It has a wide range of applications in
Pattern Recognition \cite{PARIKH2001777},
in Medical Diagnosis applications \cite{WANG20161}, in elaborating automated decision about physical security of nuclear materials \cite{Gerts22},
and in the 
Bayesian network research field \cite{7527968}.
DST was also employed in risk assessments situations, both in post-seismic structural damage and 
social vulnerability of those affected 
Three examples of newest applications are in the field of Vibration-based fault detection
\cite{Vahid22}, Sensor data fusion 
\cite{GHOSH2020113887} and
Fusion of convolutional neural networks 
\cite{Atitallah22}.

Dezert \cite{dezert2012validity} 
elaborated a pattern of entries where two sources with equal reliability and expertise are merged within DST and the resulting belief function behaves as if one of the sources was ignored. 
This is the so-called Dempster Paradox. 
The cause of this anomaly is reported to be the very structure of the DST combining process, 
called Combination Rule \cite{7194765}.

A collection of alternative combination rules was 
proposed in 
\cite{zSmarandache2006vol2}: 
the Proportional Conflict Redistribution rules. 
They share the same DST original rule principle: arithmetic over probabilistic distributions.
None of them, however, are for general purpose as each one is \textit{ad hoc} to an specific situation.

Aiming at solving this problem, we propose to replace DST Belief Fusion process (called \textit{Combination Rule}) 
by a alternative fusion method \ref{def:FusionMethod} inspired in probabilistic logic.
As far as we know, it is the first time fusion of beliefs is seen as a probabilistic logical process. 
That is the differential of our method.

        The rest of the paper is organized as follows: In Section \ref{DSTandTheProblem}, we outline the Dempster-Shafer Theory and we explain its anomaly problem. 
        In Sections 
    \ref{NovoInformationFusionInPEL} and 
    \ref{TheMergerProblem}
    we propose a solution for 
    the problem pointed out in Section \ref{DSTandTheProblem}.
    Section \ref{sec:RelatedWorkDST} focuses on related works. Finally, we sum up our work in Section \ref{sec:Conclusion}.

\section{The Dempster-Shafer Theory}
\label{DSTandTheProblem}

\subsection{Basic Definitions}\label{sec:DST21}

Dempster-Shafer Theory offers a theoretical framework for modeling uncertainty.
A finite and exhaustive collection of hypothetical answers to a question of interest 
is set
(Definition \ref{def:frameOfDiscerniment}) as well as a
weighting distribution over its power set (Definition \ref{def:BasicBeliefAssignment}).

\vspace{0.4cm}

\begin {definition} \label{def:frameOfDiscerniment} [Frame of Discernment]\cite{DST-40}. 
A \textit{frame of discernment} $\Theta$ is a finite set $\Theta = \{\theta_1, \theta_2, ..., \theta_n\}$. 
\end {definition}

Each  $\theta_i \in \Theta$ is intended to represent the following hypothesis: ``\textit {the correct value of $ \Theta $ is in $ \theta_i $}''. 
It is assumed only one hypothesis in $\Theta$ is true.  Each singleton of $\Theta$ represents a piece of information about a subject.

\vspace{0.4cm}

\begin {definition} \label{def:BasicBeliefAssignment} [Basic Belief Assignment]\cite{DST-40}:
\textit{
Let $\Theta$ be a frame of discernment. A basic belief assignment} (BBA) is a map $m$ from $2^\Theta$ to $[0, 1]$ defined as $m : 2^\theta \rightarrow{ [0,1]}$
which satisfies the conditions:
i) $\sum_{A \in 2^\Theta } m(A) = 1$ \ \ \  and \ \ \ ii) $m(\emptyset) = 0$. An element $A \in 2^\Theta$ is called a focal element if and only if $m(A) > 0$.
The vacuous BBA characterizing full ignorance is defined by $m_v(\cdot): 2^\Theta \rightarrow [0 , 1]$ such that $m_v(X) = 0$ if $X \neq \Theta$, and $m_v(\Theta) = 1$.
\end {definition}

A machinery to access the whole mass of belief in a hypothesis is given by the next definition:

\vspace{0.4cm} 

\begin {definition} \label{def:BeliefANDPlausibility function} [Belief and Plausibility functions]\cite{DST-40}: Let $\Theta$ be a frame of discernment, $m$ be a BBA and $A \in 2^\Theta$. 

$\bullet$ The Belief function is defined as
$Bel_m (A) = \sum_{\substack{
B \subseteq A}} m(B);$

$\bullet$ The Plausibility function is defined as
$Pl_m (A) = \sum_{\begin{subarray}{l}
B \cap A \neq \emptyset 
\end{subarray}} m(B)$,

\noindent where $B$ stands as a variable over $2^\Theta$.

\end {definition}

\vspace{0.4cm}

Different sources concerning the same phenomenon 
can give rise to different beliefs, formalized as BBAs over a frame of discernment. 
  Instead of choosing one belief and discarding the others, we can combine, summarize and synthesize all
 the available beliefs.

\vspace{0.5cm}

\begin{definition}\label{def:DempstersCombinationRule} [Dempster's combination rule]\cite{10.1214/aoms/1177698950}: Let $\Theta$ be a frame of discernment, $m_1$, $m_2$ be  
BBAs and $A \in 2^\Theta$. The combined BBA $m_{DS}$ is defined as $m_{DS}(\emptyset) = 0$ if $A = \emptyset$; otherwise, 
\begin{align}\label{eq:DempstersCombinationRule2}
m_{DS}(A) =  
\frac{\sum_{ A_1 \cap A_2 = A } m_1(A_1) \cdot m_2(A_2)}{1 - C(m_1 , m_2)} 
\end{align}

where $A_1$ and $A_2$ are any sets of $2^\Theta$ and

\begin{align}\label{eq:conflictFIXINGLABEL}
C(m_1 , m_2) = \sum_{ A_1 \cap A_2 = \emptyset } m_1(A_1) \cdot m_2(A_2)
\end{align}

\end {definition}

Equation \ref{eq:DempstersCombinationRule2} is the Combination Rule, and 
Equation \ref{eq:conflictFIXINGLABEL}  
is called the \textit{conflict}.
The overall BBAs mass of contradicting propositions define the conflict \cite{SHAFER201626}.
To grab this intuition, 
consider two given sets of hypothesis $X, Y \in 2^\Theta$ 
and $m_1$, $m_2$ representing the amount of \textit{belief} someone puts on those sets of hypothesis, $X$ and $Y$.
 If $X$ and $Y$ are disjoints, they code
different information, i.e., different singletons of $\theta$. 
So the product $m_1(X) \times m_2(Y)$ represents discordant \textit{beliefs} as the sets $X$ and $Y$ denotes nothing in common.  
Thus, Equation \ref{eq:conflictFIXINGLABEL} denotes a conflict of beliefs.

In Equation \ref{eq:DempstersCombinationRule2}, the numerator denotes the total amount of \textit{beliefs} put 
in $A$ while the denominator denotes the overall amount of non-conflicting belief associated with the hypothesis $A$.
Equation \ref{eq:conflictFIXINGLABEL} 
represents the proportion of conflicting belief associated with all the pairs of hypotheses from $\Omega$, i.e. $A_i$, $A_j$ 
such $A_i \cap A_j = \emptyset$.
Doing this for all elements of $2^\Theta$, we get a map from them to $[0,1]$, that is, a BBA $m_{DS}(\cdot)$ mounted over $m_1(X)$ and $m_2(Y)$. 
Example \ref{ex:DSTOverallExample} 
depicts the theory so far:

\begin{aragao}\label{ex:DSTOverallExample} (DST Overall example)

The 1990 Football World Cup had a dubious goal in the Argentine $\times$ England match. Thousands of witnesses debated between two possible hypotheses over what happened: did Maradona touch the ball with his head ($\theta_1$) or with his hand ($\theta_2$)?
Considering the frame of discernment
$\Theta = \{\theta_1, \theta_2\}$, 
the power set of possibilities is 
$2^\Theta = \{\emptyset, \ \{\theta_1\}, \ \{\theta_2\}, \ \{\theta_1, \ \theta_2\}\}$
covering all the options: $\emptyset$ meaning ``\textit{He did not touch the ball at all}'' and $\{\theta_1, \theta_2\}$ meaning ``\textit{It could have been either touched with his hand or with his head}''.
\end{aragao}

Suppose $m_1$ in which $m_1(\emptyset) = 0$, 
$m_1(\theta_1) = 1$,
$m_1(\theta_2) = 0$ and
$m_1(\{\theta_1, \theta_2\}) = 0$ is a BBA representing the opinion of an Argentine fan.
An English fan could generate the BBA $m_2$ such 
$m_2(\emptyset) = 0$, 
$m_2(\theta_1) = 0.6$, $m_2(\theta_2) = 0.3$ and
$m_2(\{\theta_1, \theta_2\}) = 0.1$.

\begin{table}[ht] \centering \begin{small}
\begin{tabular}{|c|c|c||c|}
\hline
Hypothesis    &  $m_1 (\cdot)$     &  $m_2 (\cdot)$ &  $m_{DS} (\cdot)$      \\
\hline
$\emptyset$                 & $0$                & $0$&          $0$    \\         
$\{\theta_1\}$          & $0,9$            & $0,6$ & $0,9452$  \\
$\{\theta_2\}$                 & $0$                & $0,3$ & $0,0410$  \\
$\{\theta_1, \ \theta_2\}$   & $0,1$                & $0,1$  & $0,3136$ \\         
\hline
\end{tabular}
\caption{Two BBAs $m_1$, $m_2$ and their combination, $m_{DS}$} \label{tab:Ex-Dezert2} \end{small} 
\end{table}


Each BBA $m_i$ generates a plausibility function $Pl_{m_i}(\cdot)$ and a belief function $Bel_{m_i}(\cdot)$.
The question is whether we agree with the English football fan or the Argentinian one.
In DST we resort to a combination rule (Definition \ref{def:DempstersCombinationRule}) to provide a synthesis of both.
Let calculate $m_{DS}$, the combined basic belief:
$m_{DS}(\emptyset) = 0$, 
$m_{DS}(\theta_1) = 0.56$,
$m_{DS}(\theta_2) = 0.36$ and
$m_{DS}(\{\theta_1, \theta_2\}) = 0$.

The joint belief $m_{DS}$ is different from each of the other beliefs $m_1$ and $m_2$.
This is because both sources have ``something to say'' and contributed to the final result. 
Clearly, both BBAs influence the final combination result.
Roughly speaking, this result resembles an average between the two original BBAs.

We list some virtues of DST Combination Rule: It has the commutative and associative properties and 
can aggregate evidence from different sources into a same framework.
Shafer in his seminal book \cite{shafer1976mathematical} declares: Bayes Theory is a special case of Dempster-Shafer Theory. 
In the sequel, we will present an abstract data class that causes unexpected behavior of the Combination Rule beyond the conflict incidence.

\vspace{0.5cm}

\begin {definition}\label{def:AnomalousBehaviourOfDSTCombinationRule} [Anomalous Behaviour of DST Combination Rule]\cite{dezert2012validity}: 
Let $\Theta$ be a frame of discernment and $m_1$ and $m_2$ be two BBAs (Definition \ref{def:BasicBeliefAssignment}) over $\Theta$ whith $m_1 \neq m_2$. The problem of the anomalous behaviour of DST Combination Rule (Definition \ref{def:DempstersCombinationRule}) is characterized when $m_{DS}(A) = m_1(A)$ for each $A \in \Theta$ or $m_{DS}(A) = m_2(A)$ for each $A \in \Theta$.

\end {definition}

\subsection{A Combination Rule Counter-example}\label{ACombinationRuleCounterExample}

The problem of the anomalous behaviour of DST Combination Rule (Definition \ref{def:DempstersCombinationRule}) can be characterized formally as follows:  
Think of $\Theta$ as a frame of discernment and let $m_1$ and $m_2$ be two BBAs (Definition \ref{def:BasicBeliefAssignment}) over $\Theta$. The anomaly arrives when DST Combination Rule gives the resulting BBA $m_{DS}$ replicating $m_1$ or $m_2$ exactly in all elements of $\Theta$.

DST was defended \cite{doi:10.1177/1550147717696506} 
placing the blame on the high level of conflict. However,
It is possible to create an generic input model presenting any level of conflict we want, from very low to very high, and still get an unexpected combination result.
Example \ref{ex:TheTwoDoctorsExample} takes a frame of discernment  
$\Theta = \{A, B, C\}$  
to represent distinct patient pathologies (e.g., $A$ = brain tumor, $B$ = concussion, and $C$ = meningitis):

\begin{aragao}\label{ex:TheTwoDoctorsExample}
The Two Doctors Example \cite{dezert2012validity}.

Two different doctors  
offer their individual medical diagnoses 
for the same patient based on their expertise and analyses.
The diagnostic 
corresponds to two BBA's, denoted as $m_1(:)$ and $m_2(:)$, as outlined in Table \ref{xxxx2}. The parameters $a$, $b1$, and $b2$ can assume any values, as long as $a \in [0; 1]$, $b1; b2 > 0$, and $b1 + b2 \in [0; 1]$.

\begin{table}[ht] \centering \begin{small}
\begin{tabular}{|c|c|c||c|}
\hline
Hypothesis    &  $m_1 (\cdot)$     &  $m_2 (\cdot)$ &  $m_{DS} (\cdot)$      \\
\hline
$\{A\}$    & $a$                & $0$ &          $a$    \\         
$\{A , B\}$          & $1 - a$            & $b_1$          & $1 - a$  \\
$\{C\}$                 & $0$                & $1 - b_1 - b_2$ & $0$    \\
$\{A , B , C\}$   & $0$                & $b_2$  & $0$ \\         
\hline
\end{tabular}
\caption{Two BBAs $m_1$, $m_2$ and their combination, $m_{DS}$} \label{xxxx2} \end{small} 
\end{table}
\end{aragao}

Both doctors are considered equally knowledgeable and reliable. It's essential to note that, in this parametric example 
there is a genuine conflict (Definition \ref{def:DempstersCombinationRule}, Equação \ref{eq:conflictFIXINGLABEL}) between the two sources (as demonstrated later). Also, both sources provide informative inputs, and none of them is a vacuous belief assignment representing complete ignorance. Therefore, it's reasonable to expect that both BBA's should have an impact on the fusion process.
Applying DST rule we have
$$m_{DS}(A) = \frac{a(b_1 + b_2)}{(b_1 + b_2)} = m_1(a)$$

$$m_{DS}(A \cup B) = \frac{(1 - a)(b_1 + b_2)}{(b_1 + b_2)} = 1 - a = m_1(A \cup B)$$

However, after applying the Dempster-Shafer's rule to combine the two sources of evidence, it becomes evident that, in this case, Doctor 2's medical diagnosis has no influence whatsoever, yielding $m_{DS}(\cdot) = m1(\cdot)$. Even though Doctor 2 is not an entirely ignorant source and is equally reliable as Doctor 1, their report (regardless of the values of $b_1$ and $b_2$) doesn't contribute to the result. Remarkably, 
the conflict
$C(m_1 , m_2) = 1 - b_1 - b_2$ (Equation \ref{eq:conflictFIXINGLABEL})
between the two medical diagnoses doesn't appear to matter in the DS fusion process; it can be set at high or low levels, depending on the choice of $b_1$ and $b_2$. From a DST analysis perspective, Doctor 2 appears to function as a vacuous source even though he has conflict with Doctor 1.

This outcome contradicts common sense and raises doubts about the 
usefulness of the DST rule for practical applications.  
This example 
is critical  
because 
the level of conflict between the sources does not play a role at all. Therefore, it cannot be argued that DST should not be applied in such cases due to high conflict. In reality, such situations can occur in practical applications, and the results obtained from the DST rule can have significant consequences.

Based on Zadeh's example \cite{Zadeh:M79/24} and subsequent debates in the literature, it has been widely
accepted that DST is not recommended when there is high conflict between sources. However,  
this example raises a more fundamental question because it suggests that the problem with the behavior of the DST rule is not solely due to the high level of conflict between sources; it stems from another factor. Even if we choose a low level of conflict, the result remains the same, indicating that the problem persists.

Solving this issue entails performing the fusion of the parameterized inputs and subsequently obtaining a resulting BBA that is not a replica of one of the input BBAs. This will be accomplished by altering the fusion process (Section \ref{def:FusionMethod}). This is our main objective in this work.


\section{Information Fusion}\label{NovoInformationFusionInPEL}

In Definition \ref{def:FusionMethod},
an alternative method as a replacement for the DST Combination Rule Definition \ref{def:DempstersCombinationRule} is provided
\ref{def:BBAWeightAssociation} is a auxialy definition.

\vspace{0.4cm}

\vspace{0.5cm}

\begin{definition}\label{def:BBAWeightAssociation}
		[Transformed Belief Function]: Let $ \Theta $ be a frame of discernment 
            and $2^\Theta $ be its power set and $m$ a BBA over it.
		Let 
		$ \mathcal{C} = \{v_1, ..., v_{2^n} \} $ be a collection of truth valuations 
 and 
  let $card: 2^\Theta \longrightarrow \mathbb{N}$ be a map to access the cardinality of any subset of $\Theta$.
 We define the weight function $\mu:\Theta \rightarrow [0, 1]$ 
  as   
$$\mu(\theta) = 
\sum\limits_{H \in  2^\Theta, \ \theta \in H } m(H) / card( H )$$

\end{definition}

Outlined by the DST Combination Rule Definition at \ref{def:DempstersCombinationRule}
we propose The Combined Alternative Measure at Definition \ref{def:CombinedAlternativeMeasure}, to integrates information from multiple sources.

\vspace{0.5cm}

\begin{definition}\label{def:CombinedAlternativeMeasure} [Combined Alternative Measure]: Let $\Theta$ be a frame of discernment, $m_1$, $m_2$ be 
BBAs and $A \in 2^\Theta$. The Combined Alternative Measure $m_{XXX}$ is defined as $m_{XXX}(\emptyset) = 0$ if $A = \emptyset$; otherwise, 
 
\begin{align}\label{eq:DempstersCombinationRule3}
m_{XX}(A) =  
\frac{\sum_{ A_1 \cap A_2 = A } \mu_1(A_1) \cdot \mu_2(A_2)}{K[1 - C(m_1 , m_2)]} 
\end{align}

where $A_1$ and $A_2$ are any sets of $2^\Theta$ and

\begin{align}\label{eq:conflictFIXINGLABEL2}
C(m_1 , m_2) = \sum_{ A_1 \cap A_2 = \emptyset } \mu_1(A_1) \cdot \mu_2(A_2)
\end{align}

\end {definition}

Equation \ref{eq:DempstersCombinationRule2} is the Alternative Combination Rule, Equation \ref{eq:conflictFIXINGLABEL} is called the \textit{conflict} and $K = \sum_{ 2^\Theta } \frac{\sum_{ A_1 \cap A_2 = A } \mu_1(A_1) \cdot \mu_2(A_2)}{1 - C(m_1 , m_2)}$ é uma constante de normalização.

We substitute the $m_{\text{DS}}$ BBA by applying $\mu_{XXX}$ to each element of $2^\Theta$.  
At this juncture, the merging process concludes, enabling subsequent calculations in DST without encountering the well-known anomalous outcomes. This is elaborated further in Section \ref{TheMergerProblem}.
Next, we apply this process to the imprecise information fusion issue.


\section{The Fusion Problem}\label{TheMergerProblem}

DST fusion process gave rise to the appearance of anomalous results. Several publications attribute these results to the presence of conflict in the data sets to be merged \cite{Zadeh:M79-24}.
A comprehensive list can be found here \cite{Zadeh86}. 
Dezert  \cite{dezert2012validity} identified a data pattern, according to which, the fusion results will be anomalous even in the face of low conflict, showing that the problem persists regardless of the presence of conflict or not. 
In Section 5.1 we apply our
fusion process
to the example of Section \ref{ACombinationRuleCounterExample} to show 
our main result: the alternative FusionMethod 
does not succumb to the problem elaborated by Dezert \cite{dezert2012validity}.

\subsection{A Solution to Dempster Paradox Fusion Problem}\label{secASolutionToTheFusionProblem}

 From the Definitions \ref{def:BBAWeightAssociation}, 
\ref{def:DempstersCombinationRule} we present the alternative method in Definition \ref{def:FusionMethod}

\vspace{0.4cm}

\begin{definition}\label{def:FusionMethod}
[Fusion Method]: Let $\Theta$ be a DST Frame of Discernment and let be $m_1$ and $m_2$ be BBAS over $2^\Theta$.
The combined measure $\mu_{XX}$
from $m_1$ and $m_2$ is defined as
\begin{description}
    \item[A] Convert $m_1$ and $m_2$ into measures $\mu_1$ and $\mu_2$ by Definition \ref{def:BBAWeightAssociation}.
    \item[B] Unify the $\mu_1$ and $\mu_2$ using the Combined Alternative Measure
    \ref{def:DempstersCombinationRule} getting $\mu_{XX}$
    \end{description} 
  $\mu_{XX}$ is the transcription of Dempster’s combination rule $m_{DS}$ in Definition \ref{def:DempstersCombinationRule}.
    
\end{definition}

\vspace{0.4cm}

Next, we apply this method to the Two Doctors Problem seen from Section \ref{ACombinationRuleCounterExample}. 
And calculate the subsequent the Combined Alternative Measure of hypothesis $ {H_3} $, the crux of the example.

\begin{aragao}\label{Ex:merging}(Merging): 

First, we calculate $\mu_1(C)$ and $\mu_2(C)$ using Definition \ref{def:BBAWeightAssociation}:

$$\mu_1(\{C\}) = m_1(\{C\}) / 1 + m_1(\{A, C\}) / 2 +
m_1(\{B, C\}) / 2 +
m_1(\{A, B, C\}) / 3$$

According to the values for $m_1$ and $m_2$ in Table \ref{xxxx2}, we obtain $\mu_1(\{C\}) = 0$.

$$\mu_2(\{C\}) = m_2(\{C\}) / 1 + m_2(\{A, C\}) / 2 +
m_2(\{B, C\}) / 2 +
m_2(\{A, B, C\}) / 3$$

Based on the values in Table \ref{xxxx2}, we obtain 
$
\mu_2(\{C\}) = (1 - b_1 - b_2 + b_2/3)
$.

We calculate $m_{XX}(C)$ using Definition \ref{def:DempstersCombinationRule}:

{\small 
\begin{align*}
m_{XX}(C) = \frac{1}{K\left[1 - C(m_1, m_2)\right]} \Bigg( \mu_1(C) \cdot \mu_2(C) + \mu_1(C) \cdot \mu_2(A, C) + \\ \mu_1(C) \cdot \mu_2(B, C) + \mu_1(C) \cdot \mu_2(A, B, C) + \mu_1(A, C) \cdot \mu_2(C) +
\mu_1(B, C) \cdot \mu_2(C) + \mu_1(A, B, C) \cdot \mu_2(C) \Bigg)
\end{align*}
}

that is, $m_{XX}(C) = 0$.

Observe that the constants $K$ and $C(m_1, m_2)$ are calculated as in Definition \ref{def:DempstersCombinationRule}. 

\end{aragao}

The Fusion Method \ref{def:FusionMethod} allows us to calculate the Combined Logical Measure $\mu_{XX}$ we look for to replace DST Combination Rule $m_{DS}$, see Table 
\ref{xxxx3}.

\begin{table}[ht] \centering \begin{small}
\begin{tabular}{|c|c|c||c|}
\hline
Hypothesis    &  $m_1 (\cdot)$     &  $m_2 (\cdot)$ &  $m_{XX} (\cdot)$      \\
\hline
$\{A\}$   & $a$           & $0$ &         $\neq a$ \\ 
$\{A , B\}$          & $1 - a$            & $b_1$          & $1 - a$  \\
$\{B\}$                & $0$                & $0$ &          $\neq 0$  \\  
$\{C\}$                 & $0$                & $1 - b_1 - b_2$ & $0$    \\
\hline         
$\{A , B , C\}$   & $0$                & $b_2$  & $0$ \\         \hline
\end{tabular}
\caption{Two BBAs $m_1$, $m_2$ and their combination, $m_{XX}$} \label{xxxx3} \end{small} 
\end{table}

We fusion the beliefs $m_1$ and $m_2$ using the  
Combined Alternative Measure (Definition \ref{def:DempstersCombinationRule}
).  
The obtained distribution represents the merged BBA we are looking for.
We found that $\mu_{XX}(C)$ does not present the DST problems $m_{DS} (\cdot)$ presents. I.e. The Combined Alternative Measure can fuse the imprecise information of the ``Two Doctors Problem'' without falling into contradictory conclusions.

The problematic combined BBA $m_{DS}$ from Table \ref{xxxx2} will be replaced by a combined New Measure. 
The focal elements (Definition \ref{def:BasicBeliefAssignment})  
measurements are in Table \ref{xxxx2},
last column. The Combined Alternative Measure 
of hipothesis $ \{C\} $ despite been null, do note neclects Doctor two's opinion. More on that issue, see \cite{dezert2012validity}.

Remark 1: 
DST Combination Rule leads to $m_{DS}(C) = 0$ and this opinion cannot be changed in the light of new evidence. 
For the same pattern, The combined New Measure evaluates the third hypothesis as been zero, $ \mu_{XX} (C) = 0 $ but $ \mu_{XX} \neq m_1 $. This result is stated in Theorem \ref{teo:WhatsAppBIS}.

\subsection{ Conclusion  of Two Doctors Example}

DST produces inappropriate conclusions 
in merging information. 
Those problematic issues disapears due to the fusion process based  
on an isomorphism between DST and a probabilistic logic.

$ \bullet $ In DST, the combined opinion $ m_{DS} $ is the same as that of Doctor 1 $ m_1 $. This is a DST flaw pointed out in this example.
We managed to show that the new measurement, $\mu_{XX}$ is different from Doctor 1's distribution $ m_1 $ seen in Table 
\ref{xxxx2}.

$ \bullet $ In DST, $ m_{DS} (C) = 0 $, which is an inadequate result as it cancels the Doctor 2 opinion. We show that the new measurement of $ \mathcal{C} $, even if it is zero for the inputs in question, does not cause an anomalous result.

\section{Results and Conclusion}\label{sec:Teorema}

The combined $\mu_{XX}$ measure corresponding to the DST combined BBA $m_{DS}$ is obtained by applying the new fusion method Definition \ref{def:FusionMethod}.
Theorem \ref{teo:WhatsAppBIS} shows how these concepts are connected solving the Two Doctors Problem Section \ref{ACombinationRuleCounterExample}.

Theorem \ref{teo:WhatsAppBIS} shows that [Combined Alternative Measure], Definition \ref{def:CombinedAlternativeMeasure} combine belief functions from Dezert template without producing anomalies.

\begin{theorem}\label{teo:WhatsAppBIS}
Let $\Theta$ be a frame of discernment and $m_1$ and $m_2$ be two BBA over $\Theta$. Let $\mu_1$ and $\mu_2$ be transcripted measures of $m_1$ and $m_2$. There exists $A \in \Theta$ such that $\mu_{XX}(A) \neq m_1(A)$ and $\mu_{XX}(A) \neq m_2(A)$.
\noindent Where $\mu_{XX}(\cdot)$ is the combined measure of $\mu_1(\cdot)$ and $\mu_2(\cdot)$ by the Combined Alternative Measure (Definition 
\ref{def:CombinedAlternativeMeasure}).
\end{theorem}

To prove Theorem \ref{teo:WhatsAppBIS}, the following lemma on BBAs differentiation is required.

\begin{lemma}\label{teo:Lema01julhoC}
Let \(m_1\) and \(m_2\) be two BBAs of the same size \(n\) whose contents are in \([0, 1]\) and whose respective summations result in 1.

if \(m_1 \neq m_2\), then there exist at least \(k \in \{1, ..., n\}\) and \(l \in \{1, ..., n\}\) such that 
\(m_1(k) \neq m_2(k)\)
and
\(m_1(l) \neq m_2(l)\).
\end{lemma}

\begin{aragaoProva}

Given that \( m_1 \neq m_2 \), there exists an index \( i \in \{1, \ldots, n\} \) such that \( m_1(i) \neq m_2(i) \). Let us denote this index by \( i_1 \). That is,
\[ m_1(i_1) \neq m_2(i_1). \]

We need to show that there exists a second index \( j \) 
(different from \( i_1 \)) such that \( m_1(j) \neq m_2(j) \).
Suppose, for contradiction, that \( m_1(i) = m_2(i) \) for all \( i \neq i_1 \):
\[ m_1(i) = m_2(i) \quad \text{for all } i \in \{1, \ldots, n\} \setminus \{i_1\}. \]

By the Theorem \ref{teo:WhatsAppBIS} hypothesis we have:
\[ \sum_{i=1}^{n} m_1(i) = 1 \quad \text{and} \quad \sum_{i=1}^{n} m_2(i) = 1. \]

Thus:
\[ \sum_{\substack{i=1 \\ i \neq i_1}}^{n} m_1(i) + m_1(i_1) = 1 \quad \text{and} \quad \sum_{\substack{i=1 \\ i \neq i_1}}^{n} m_2(i) + m_2(i_1) = 1. \]

By the hypothesis of contradiction, we assumed that \( m_1(i) = m_2(i) \) for all \( i \neq i_1 \), which implies that:
\[ \sum_{\substack{i=1 \\ i \neq i_1}}^{n} m_1(i) = \sum_{\substack{i=1 \\ i \neq i_1}}^{n} m_2(i). \]

Subtracting this equality from the above equations, we get:
\[ m_1(i_1) = 1 - \sum_{\substack{i=1 \\ i \neq i_1}}^{n} m_1(i) \quad \text{and} \quad m_2(i_1) = 1 - \sum_{\substack{i=1 \\ i \neq i_1}}^{n} m_2(i). \]

Therefore,
\[ m_1(i_1) = 1 - \sum_{\substack{i=1 \\ i \neq i_1}}^{n} m_1(i) = 1 - \sum_{\substack{i=1 \\ i \neq i_1}}^{n} m_2(i) = m_2(i_1). \]

This contradicts the initial assumption that \( m_1(i_1) \neq m_2(i_1) \). 
Hence, there is at least one other index \( j \) such that \( m_1(j) \neq m_2(j) \).

\end{aragaoProva}

\subsection{Proof outline of the theorem}

\begin{enumerate}
\item The hypothesis of contradiction leads to a system of equations, Section  \ref{sec:SitemaDeEquações}.
\item The hypothesis of the theorem leads to Lemma \ref{teo:Lema01julhoC} on inequalities in BBAs.
\item The Lemma \ref{teo:Lema01julhoC} and the constraints of the BBA definition lead to the inconsistency of the system of equations.  
\item The system without a solution in the real interval \([0, 1]\) constitutes a contradiction to the hypothesis of contradiction.
\end{enumerate}

\noindent Items 2 and 3 constitute the technical part of the proof.

\subsubsection{The hypothesis of contradiction leads to a system of equations}\label{sec:SitemaDeEquações}

Looking at Definition \ref{def:CombinedAlternativeMeasure} of Combined Alternative Measure 
and at the hypothesis of contradiction\footnote{$m_1(A) = m_{XX}(A)$ for all $A \in 2^\theta$.}, we have that, for all $A \in 2^\Theta$

$$m_1(A) = \frac{1}{K^*}  
\sum_{ A_1 \cap A_2 = A } \big(\sum\limits_{H \in  2^\Theta, \ A_1 \in H } m_1(H) / card( H ) \cdot \sum\limits_{G \in  2^\Theta, \ A_2 \in G } m_2(G) / card( G )\big) $$

$\Rightarrow$

\begin{align}\label{eqsistema}
K^* = \frac{
\sum_{ A_1 \cap A_2 = A } \big(\sum\limits_{H \in  2^\Theta, \ A_1 \in H } m_1(H) / card( H ) \cdot \sum\limits_{G \in  2^\Theta, \ A_2 \in G } m_2(G) / card( G )\big)}{m_1(A)}
\end{align}

This expression generates a system of nonlinear equations of size $2^n$, where all the equations are equal to \(K^*\).

By definition \ref{def:CombinedAlternativeMeasure}, \(m_{XX}\) \(K^*\) is a normalization constant, so it must be unique, which does not happen because, by the hypothesis of the theorem, and by Lemma 1, there exist \(k, l \in 2^\Theta\) such that \(m_1(k) \neq m_2(k)\) and \(m_1(l) \neq m_2(l)\).
Adding the summation equality and the inequalities constraints of the BBA Definition, the aforementioned system has no real solution in \([0, 1]\), which constitutes a contradiction to the hypothesis of contradiction.

\subsubsection{The theorem hypothesis leads to Lemma \ref{teo:Lema01julhoC} on inequalities in BBAs}

\subsubsection{Lemma \ref{teo:Lema01julhoC}, Lemma \ref{teo:Lema01julhoB} and the constraints of the BBA definition lead to the inconsistency of the system of equations}
\subsubsection*{Proof by contradiction and induction}
Denying the thesis of Theorem \ref{teo:WhatsAppBIS} we have the following hypothesis of contradiction:
\begin{align}\label{hyp:absurd}
m_1(A) = m_{XX}(A) \textrm{ for all } A \in 2^\theta
\end{align}

We show that Equation \ref{hyp:absurd} generates a contradiction by induction on the size of the 'Frame of discernment' Definition \ref{def:frameOfDiscerniment}, \(\Theta\).

\noindent 
\noindent \textbf{\large Property:} \textit{The system generated by the hypothesis of contradiction of Equation \ref{hyp:absurd} is inconsistent.}

\noindent \textbf{Base case:} Check for \(n = 2\), we have \(\Theta = \{\theta_1, \theta_2 \}\)

Mostramos que a equação \ref{hyp:absurd} gera um absurdo por indução no tamanho do `Frame of discernment' Definição \ref{def:frameOfDiscerniment}, $\Theta$.

By Lemma \ref{teo:Lema01julhoC},
by definition \ref{def:CombinedAlternativeMeasure}, and by the constraints of Theorem \ref{teo:WhatsAppBIS}, we have 

1. For \(A = \{\theta_1, \theta_2\}\):
\[
K^* = \frac{m_1(\{\theta_1, \theta_2\}) \cdot m_2(\{\theta_1, \theta_2\})}{4 m_1(\{\theta_1, \theta_2\})} = \frac{m_2(\{\theta_1, \theta_2\})}{4}
\]

2. For \(A = \{\theta_1\}\):
\[
K^* = \frac{\left( m_1(\{\theta_1\}) + \frac{m_1(\{\theta_1, \theta_2\})}{2} \right) \cdot \left( m_2(\{\theta_1\}) + \frac{m_2(\{\theta_1, \theta_2\})}{2} \right)}{m_1(\{\theta_1\})}
\]

3. For \(A = \{\theta_2\}\):
\[
K^* = \frac{\left( m_1(\{\theta_2\}) + \frac{m_1(\{\theta_1, \theta_2\})}{2} \right) \cdot \left( m_2(\{\theta_2\}) + \frac{m_2(\{\theta_1, \theta_2\})}{2} \right)}{m_1(\{\theta_2\})}
\]

By substituting the functional notation with variables \(x_i\), \(y_i\), and \(z_i\), and equating the above equations, we obtain the equations

\[
\frac{z_2}{4}
= \frac{\left( x_1 + \frac{z_1}{2} \right) \left( x_2 + \frac{z_2}{2} \right)}{x_1}
\]

\[
\frac{\left( x_1 + \frac{z_1}{2} \right) \left( x_2 + \frac{z_2}{2} \right)}{x_1}
= \frac{\left( y_1 + \frac{z_1}{2} \right) \left( y_2 + \frac{z_2}{2} \right)}{y_1}
\]

\[
\frac{z_2}{4}
= \frac{\left( y_1 + \frac{z_1}{2} \right) \left( y_2 + \frac{z_2}{2} \right)}{y_1}
\]

\noindent which, together with the constraints of the theorem, form a nonlinear system without a real solution in the interval \([0, 1]\).

\noindent \textbf{Inductive step:}

Assume 
the property holds for \(n = k\), i.e., assume the hypothesis of contradiction generates an inconsistent system for \(\Theta\) with \(k\) elements.
By Lemma \ref{teo:Lema01julhoB}, adding an additional element to the set\footnote{$\Theta$} does not change the inconsistency of the system\footnote{engendered by $\Theta' = \{\Theta \cup \theta{n + 1}\}$,}  as the system\footnote{engendered by $\Theta$,} is fundamentally inconsistent for any \(k\)-sized subset\footnote{and the restrictions from \ref{hyp:absurd} still applies to the new system.}\footnote{here there would be some technicalities that were skipped.}. Therefore, the system remains inconsistent for \(k+1\) elements, proving that the hypothesis of contradiction leads to an inconsistent system for \(n = k+1\).

This completes the induction and we conclude that the hypothesis of contradiction generates an inconsistent system for all \(n \geq 2\).
Since the hypothesis of contradiction leads to inconsistency, the original thesis of Theorem \ref{teo:WhatsAppBIS} must be true.
\qed

This expression generates a system of nonlinear equations of size $2^n$, where all the equations are equal to \(K^*\).

By definition \ref{def:CombinedAlternativeMeasure}, \(m_{XX}\) \(K^*\) is a normalization constant, so it must be unique, which does not happen because, by the hypothesis of the theorem, and by Lemma 1, there exist \(k, l \in 2^\Theta\) such that \(m_1(k) \neq m_2(k)\) and \(m_1(l) \neq m_2(l)\).

Adding the summation equality and the inequalities constraints of the BBA Definition (\ref{def:BasicBeliefAssignment}), the aforementioned system has no real solution in \([0, 1]\), which constitutes a contradiction to the hypothesis of contradiction.

\subsubsection{The system without a solution in the real interval \([0, 1]\) constitutes a contradiction to the hypothesis of contradiction.}

We've shown that an absurd hypothesis leads to an absurdity, so the theorem is proven.

\begin{lemma}\label{teo:Lema01julhoB}
Let $x_i$ and $y_i$ be two sets of variables contained in the real interval $[0, 1]$ such that $i \in \{1, ..., n\}$ and $\sum_{i=1}^{n} x_i = 1$ and $\sum_{i=1}^{n} y_i = 1$. 
Let $k, l \in \{1, ..., n\}$ such that $x_k \neq y_k$.
Let there be a nonlinear system of $n$ equations without a solution in the real interval $[0,1]$.
Adding the variables $x_{n+1}$ and $y_{n+1}$ to the respective sets of variables $x_i$ and $y_i$ 
does not alter the inconsistency of the new system created by the new sets of variables.

\end{lemma}

\begin{aragaoProva}

$\bullet$ Initial Variables 

   We have sets of variables \( \{x_i\} \) and \( \{y_i\} \) where \( i \in \{1, ..., n\} \), all within the interval \([0, 1]\) such as.

  \[
  \sum_{i=1}^{n} x_i = 1 \quad \text{and} \quad \sum_{i=1}^{n} y_i = 1.
  \]
   
   There exist \( k \) and \( l \) such that \( x_k \neq y_k \) and \( x_l \neq y_l \) with \( k \neq l \).

   The original system of \( n \) nonlinear equations in \( \{x_i\} \) and \( \{y_i\} \) is inconsistent.

We add \( x_{n+1} \) to the set \( \{x_i\} \) and \( y_{n+1} \) to the set \( \{y_i\} \).
- The new sums are:
  \[
  \sum_{i=1}^{n+1} x_i = 1 \quad \text{and} \quad \sum_{i=1}^{n+1} y_i = 1.
  \]

If we add positive terms with the new variables \( x_{n+1} \) and \( y_{n+1} \) in the equations, the system expands, but we still need to check for consistency.

   The sums now include \( x_{n+1} \) and \( y_{n+1} \):
     \[
     \sum_{i=1}^{n} x_i + x_{n+1} = 1 \quad \text{and} \quad \sum_{i=1}^{n} y_i + y_{n+1} = 1.
     \]
   Since \(\sum_{i=1}^{n} x_i = 1 - x_{n+1}\) and \(\sum_{i=1}^{n} y_i = 1 - y_{n+1}\), \( x_{n+1} \) and \( y_{n+1} \) can vary within the interval \([0, 1]\), but they still must satisfy the new equations.

   Suppose the new equations have the form \( f_i(\{x_j\}, x_{n+1}) \) and \( g_i(\{y_j\}, y_{n+1}) \) where \( i \in \{1, ..., n+1\} \).
   The original system was inconsistent, meaning that there were no values \( \{x_i\} \) and \( \{y_i\} \) that satisfied all the equations simultaneously.

Adding new equations, we are still constrained by the normalization conditions and the original inconsistency. Even with \( x_{n+1} \) and \( y_{n+1} \) added, the original inconsistency implies that there is no combination of \( \{x_i\}, x_{n+1} \) and \( \{y_i\}, y_{n+1} \) that satisfies all the equations simultaneously.

Therefore, the new system with \( n+1 \) variables and additional equations, even with positive terms involving \( x_{n+1} \) and \( y_{n+1} \), remains inconsistent due to the inherent inconsistency of the original system. The introduction of new variables and terms does not resolve the original system fundamental inconsistency.
\end{aragaoProva}

\section{Related Work - DST-CR}\label{sec:RelatedWorkDST}

Ever since the issuing of Zadeh’s example in \cite{Zadeh:M79/24},
    Dempster’s rule has been criticized intensively, mainly in the presence of rather conflicting beliefs due to using his Combination Rule and getting counterintuitive results \cite{6335122}. 
Many tries to modify Dempester’s Combination Rule were made to bridle this conflict. Those modifications  
may be classified into 2 types. 
The idea in the first type of techniques is to transfer overall or partial conflicting masses proportionally to empty or non-empty sets according to a few
combination outcomes. 
The second type of strategies are based totally on the essential principle of Dempster’s rule and, in addition, they use a correction to original evidences \cite{10.1016/S0167-9236(99)00084-6,YAMADA20081689,chena2011more,inproceedingsX,YANG20131}.
These strategies are based on mathematics average, a distance measure or merging of the evidences to be combined. 
Nevertheless, those classes of rules have a few flaws. 
\cite{6916059} cites another instance of misbehavior related to the absence of conflicting beliefs.

\subsubsection*{Type 1 approach}

Type 1 approaches attempts to overcome the counterintuitive behaviors by modifying the Dempster’s rule to solve the problem of how to redistribute and manage evidential conflict. They differ from Dempster’s rule approach.

Because the conflict cannot provide any useful information, Yager proposed a modified rule which removes the normalizing process of Dempsters rule. 
The proposed combination rule also has its shortcomings, as it assigns conflicting mass assignments to the universal set of the frame of discernment and is not associative.

The Dubois and Prade rule \cite{Dubois1988RepresentationAC} is based on the disjunctive operator and applies the opposing mass assignment to the union of focal elements, forcing us to choose between potential combinations. However, this disjunctive rule frequently reduces data specificity. 

Smets \cite{55104} proposal is equivalent to Dempster's rule in its non normalized form. Using the close-world assumption, this technique allocates the conflicting mass assignments to an empty set. Those named Lefevre et al. 

\cite{LEFEVRE2002149} presented a unified formulation for combination rules based on the conjunctive agreement and the reallocation (convex combination) of conflicting masses.

According to various constraints, the goal behind these rules is to transmit total or partial conflicting masses proportionally to the non-empty set and the partial ignorance involved in the model.

Dezert and Smarandache \cite{Smarandache2006vol3} proposed a number of proportional conflict redistribution methods (PCR 1 to 6) to redistribute the partial or total conflicting mass on the focal factors included in the local conflict. 
The authors demonstrated that these rules do not provide surprising conclusions in strongly contradictory situations, but they are not associative, and their implementation is more difficult than Dempster's rule \cite{inproceedingsY}. 
As a consensus generator, an alternate combination process known as combination by compromise \cite{YAMADA20081689} can be used.

\subsubsection*{Type 2 approach}

The second category 
is based on the remedial procedures when utilizing Dempster’s
run the show.
Murphy proposed a combination rule \cite{10.1016/S0167-9236(99)00084-6} based on arithmetic average of belief functions related with the
evidences to be combined. This can be a commutative but not associative trade-off rule and tackles the vanishing
ignorance issue caused by the operation of intersection in Dempster’s rule.
In any case, this strategy does not consider the association relationship between the evidences, which isn't sensible in a few circumstances.

Following that, Deng et al. \cite{chena2011more} took the same notion as \cite{articleYong}, and provided a more effective definition of evidentiary conflict based on the evidential distance measure. However, when compared to the first category of methods, the prior methods based on distance measure are applicable to a finite frame of discernment and can improve the dependability of the combination outcomes. 

In \cite{YANG20131}, another strategy called the Weight Evidential Rule is proposed to generate a new Weighted Belief Distribution to be coupled.

\subsubsection*{Disadvantages of both}

These two types of rules have some drawbacks. The first set of guidelines assumes that the sources are contradictory and favors the sources with the most evidence in the conflict redistribution stage. When this is not the case, these rules behave similarly to Dempster's rule.

To some extent, the problem of the second class of rules can be solved by substituting the evidences with improperly small weighted evidence values generated using distance functions. 
As a result, these distances do not appear to be as well justified and do not make use of all the facts bodies rest on \cite{articleGuo}. 
Furthermore, these criteria do not prevent the masses of focal elements from possessing a low level of specific information (partial or entire ignorance) in the combination process.
The reason is that Dempster rule and its derivative assume that all possible pairs of focal elements are equally confirmed by the combined evidence  
\cite{VOORBRAAK1991171}.

\section{Conclusion}\label{sec:Conclusion}

While the problem with DST Combination Rule is by now common knowledge, alternative rules like those in Section \ref{sec:RelatedWorkDST}
continually fail to solve this anomaly topic due to Dempster Paradox.

The common cause rests in the algebraic nature of those rules. To create a truly alternative information fusion, we suggest transitioning from algebraic tools to a logic probabilistic procedure. This way, we can achieve an imprecise-data fusion robust to conflict between sources, as well as
free from Dempster Paradox trap. Although this approach would have complexity issues from a theoretical perspective, it would represent a significant step towards a more efficient imprecise-data fusion procedures.
procedure will make the merge process broader and safer, avoiding the Dempster Paradox problem.


\end{document}